\pdfoutput=1
\documentclass[11pt]{article}

\usepackage{acl}

\usepackage{times}
\usepackage{latexsym}
\usepackage{multirow}   %  table
\usepackage{graphicx}   %  .png
\usepackage{enumitem}
\usepackage{amsmath}    %  symbols
\usepackage{amssymb}    %  symbols
\usepackage{latexsym}
\usepackage[inkscapelatex=false]{svg}
\usepackage{url}    % \url
\usepackage{colortbl, booktabs} % To thicken table lines
\usepackage{makecell} % automatic words wrap
\usepackage{color}  %  color
\usepackage{array}
\usepackage{hyperref}   %  \href{url}{caption}
\usepackage[all]{nowidow}   % prevents the appearance of an ugly single line on the second page.
\usepackage[subtle]{savetrees} % The goal of the savetrees package is to pack as much text as possible onto each page

\usepackage{algorithm}
\usepackage{algorithmic}

\usepackage{microtype}
\usepackage{multicol}
\usepackage{tipa}

\usepackage{inconsolata}

\usepackage[T1]{fontenc}

\usepackage[utf8]{inputenc}

\title{With~Ears~to~See~and~Eyes~to~Hear: Sound Symbolism Experiments with Multimodal Large Language Models}

\author{Tyler Loakman\textsuperscript{1}, Yucheng Li\textsuperscript{2}  ~and Chenghua Lin\textsuperscript{1, 3 \thanks{Corresponding author}}\\
  \textsuperscript{1}Department of Computer Science, The University of Sheffield, UK \\
  \textsuperscript{2}Department of Computer Science, University of Surrey, UK \\
  \textsuperscript{3}Department of Computer Science, The University of Manchester, UK\\
  \texttt{tcloakman1@sheffield.ac.uk} \\
    \texttt{yucheng.li@surrey.ac.uk}~~~~
    \texttt{chenghua.lin@manchester.ac.uk}}

\begin{document}

\maketitle

\begin{abstract}
Recently, Large Language Models (LLMs) and Vision Language Models (VLMs) have demonstrated aptitude as potential substitutes for human participants in experiments testing psycholinguistic phenomena. However, an understudied question is to what extent models that only have access to vision and text modalities are able to implicitly understand sound-based phenomena via abstract reasoning from orthography and imagery alone. To investigate this, we analyse the ability of VLMs and LLMs to demonstrate sound symbolism (i.e., to recognise a non-arbitrary link between sounds and concepts) as well as their ability to ``hear'' via the interplay of the language and vision modules of open and closed-source multimodal models. We perform multiple experiments, including replicating the classic Kiki-Bouba and Mil-Mal shape and magnitude symbolism tasks and comparing human judgements of linguistic iconicity with that of LLMs. Our results show that VLMs demonstrate varying levels of agreement with human labels, and more task information may be required for VLMs versus their human counterparts for \textit{in silico} experimentation. We additionally see through higher maximum agreement levels that Magnitude Symbolism is an easier pattern for VLMs to identify than Shape Symbolism, and that an understanding of linguistic iconicity is highly dependent on model size.
\end{abstract}

\section{Introduction}

\begin{figure}
    \centering
    \includegraphics[width=1\linewidth]{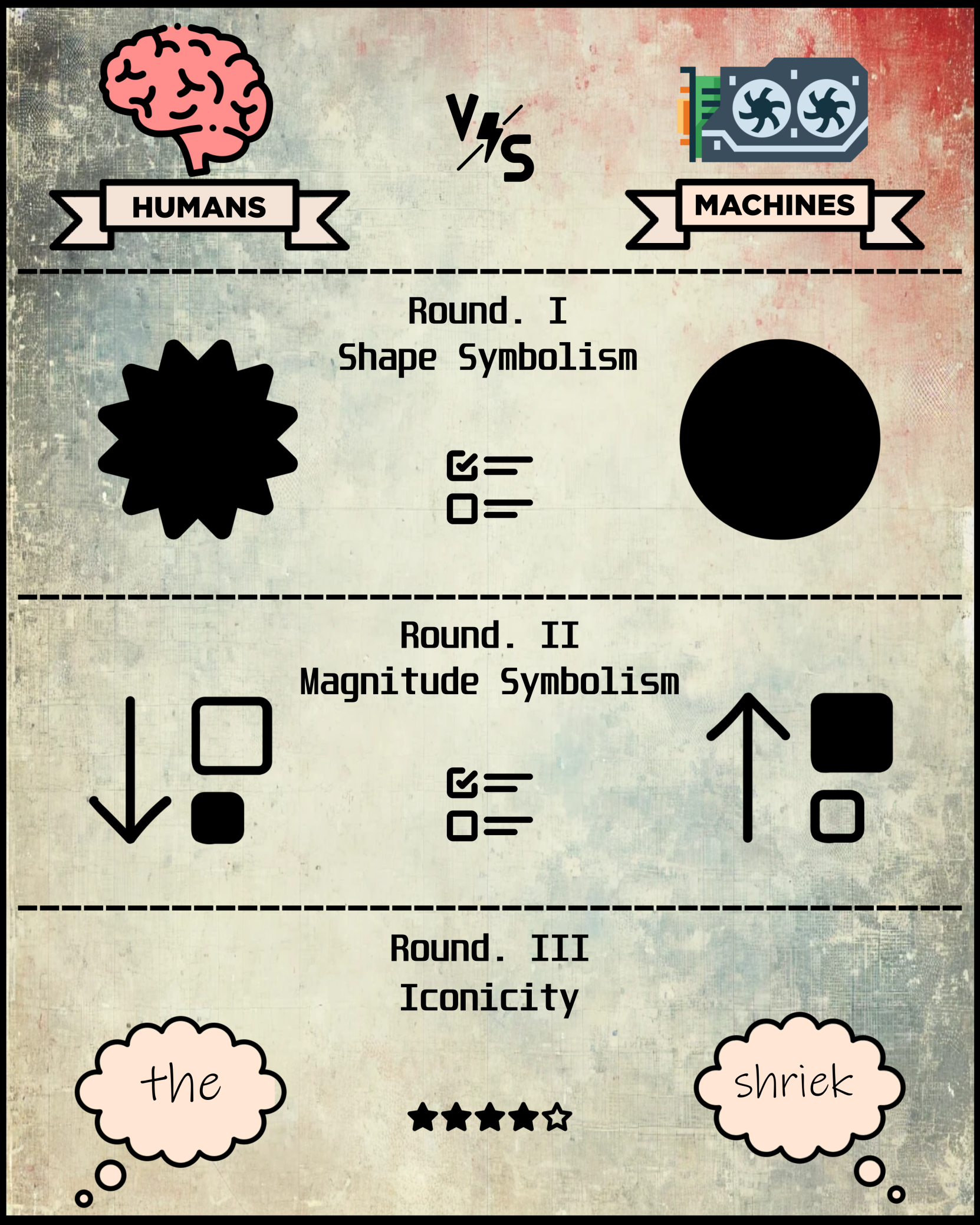}
    \caption{Illustration of the 3 main experiments we perform. Firstly, Shape Symbolism is a binary choice between two pseudowords to best describe an object that is spiky or rounded. Magnitude Symbolism involves a binary choice between two pseudowords to best describe an object that is small or large. Finally, Iconicity involves rating the perceived iconicity of words, or how much their written/phonetic form is representative of what they describe.}
    \label{fig:first-page}
\end{figure}

Sound symbolism refers to a perceived similarity between speech sounds and the conceptual meanings of the words they comprise. Evidence of this can be found in linguistic devices such as onomatopoeia (e.g., ``bang'', ``shriek'', and ``bellow''), where a word imitates the concept it describes via its phonetic form. The ability of Large Language Models (LLMs) and Vision Language Models (VLMs) to reflect a sense of sound symbolism would therefore suggest that these models are capable of acquiring ``phonetic'' knowledge indirectly through only the written orthographic form of a language via patterns of grapheme combinations \cite{loakman2024,loakman-etal-2023-twistlist} and meta-level textual discussion of sound in training data, which has implications for the potential future use of LLM/VLMs in perceptual studies usually reserved for humans \citep{jain-2023-silico, DILLION2023, aher2023}. In this work, we further explore the capability of LLMs/VLMs to demonstrate human-like characteristics in a range of psycholinguistic perceptual tests investigating 3 main areas of sound symbolism: (1) \textit{Shape Symbolism} (i.e., the Kiki-Bouba effect, \citealp{Ramachandran2001}), where a forced choice must be made between two pseudowords as to which is the most appropriate to describe shapes and entities that are spiky or rounded; (2) \textit{Magnitude Symbolism} (i.e., the Mil-Mal Effect, \citealp{sapir-1929}), a similar test to (1), but where the entities are small or large (rather than spiky or rounded); and (3) \textit{Iconicity Rating} \cite{Winter2023}, where LLMs are asked to rate a series of English words on their perceived ``iconicity'' (i.e., to what extent a word's form is perceived to be analogous to the concept or entity it describes). We illustrate these experiments in \autoref{fig:first-page}.

By extending these experiments to LLMs/VLMs, our study aims to shed light on the processes underlying multimodal perception in language models.\footnote{Our paper title is inspired by the Sleeping With Sirens album of the same name: \url{https://en.wikipedia.org/wiki/With_Ears_to_See_and_Eyes_to_Hear}.} Moreover, the presence of sound symbolism in such models could inform the development of more effective natural language processing algorithms, aiding tasks such as sentiment analysis, emotion recognition, and content generation that take into account more abstract layers of human reasoning and perception such as abstract connotations between words rather than semantics alone \cite{manzoor-multimodal}. An understanding of sound symbolism also has the potential to have a profound effect on creative generation, including language forms such as poetry and narratives (and their accompanying illustrations). Additionally, sound symbolism is a prevalent strategy used in marketing products to create desirable associations in potential customers, and LLMs capable of understanding this phenomenon could be used as pilot testing before using focus groups to reduce time and monetary costs \citep{KetronSeth2021Ssoo, MotokiKosuke2020TnSi, SpenceCharles2012Msec}.

We summarise our main contributions as follows:\footnote{We release our code and resources on GitHub: \url{https://github.com/tylerL404/WETSAETH/}.}
\begin{itemize}
    \item We perform replications of the classic psycholinguistic Kiki-Bouba Shape Symbolism and Mil-Mal Magnitude Symbolism studies with a range of open and closed-source VLMs to investigate if they understand the association between speech sounds/orthographic forms and the characteristics of entities.
    \item We perform an in-depth analysis of the ability of a range of closed and open-source LLMs to demonstrate an understanding of linguistic iconicity by comparing judgements to an existing large-scale dataset of human ratings.
    \item We provide a discussion of the potential sources of sound symbolism abilities in LLM/VLMs and potential future approaches to bolstering these abilities, in addition to the implications of doing so in \S\ref{sec:discussion}.
\end{itemize}

%Through these contributions, we are able to address the following research questions (\textbf{RQs}):
%\begin{itemize}[noitemsep]%[leftmargin=*]
%    \item \textbf{RQ1:} Do multimodal LLMs demonstrate human-like preferences regarding Shape Symbolism as demonstrated by the classic Kiki-Bouba test?
%    \item \textbf{RQ2:} Do multimodal LLMs demonstrate human-like preferences regarding Magnitude Symbolism as demonstrated by the classic Mil-Mal test?
%    \item \textbf{RQ3:} To what extent do multimodal LLMs' ratings of linguistic iconicity correlate with those of humans?
%\end{itemize}

\section{Related Works} 
The work of early linguists, such as Saussure, touched upon the topic of whether or not the link between the "sign" (i.e., a word) and the "signified" (i.e., the entity/concept to which the sign relates) is arbitrary, with there being nothing more ``boat''-esque about the word ``boat'' than any other combination of phonotactically legal sounds \citep{saussure-general}. However, there are many types of language where this association is seen to be \textit{non}-arbitrary such as in the onomatopoeia commonly used in literary works (e.g., ``bang'' for a loud noise, or ``shriek'' for a high-pitched wail), where the phonetic realisation mirrors the concept it describes.
These phenomena as a whole are known as sound symbolism, where there is thought to be a non-arbitrary link between the sign and the signified, in contrast with the popular stance of early linguistics. 

Outside of onomatopoeia, sound symbolism is believed to have a range of effects on human perception, including applying to nonsense pseudowords. For example, even if a word was not created to explicitly denote a known concept or entity (and therefore has no true \textit{denotative} meaning), it is nevertheless able to manifest a \textit{connotative} meaning in the mind of the reader based on its phonological representation and/or phonetic realisation. 
The first identification of these patterns is frequently attributed to \citet{usnadze_1924}, who gave 10 participants a series of pseudowords alongside drawings and found a higher-than-chance level of agreement between evaluators for which nonsense word best described which drawing. Perhaps the most famous example of this is in the Kiki-Bouba effect \citep{Sidhu2021, Ramachandran2001,kohler-1929-takete} which concerns the allocation of the name ``Kiki'' to sharp, hard-edged entities, and ``Bouba'' to more soft and round-edged entities (which in the original works consisted exclusively of 2D shapes). 
A similar relationship has also been noted between the words ``Mil'' and ``Mal'', where the changing vowel in the phonological minimal pairs\footnote{A \textit{Minimal Pair} refers to a pair of words that differ only in one phonological segment, such as ``cat'' \textipa{/kat/} versus ``bat'' \textipa{/bat/}.} has a relationship to perceived size, in a phenomenon known as magnitude symbolism (where vowels with higher frequency content are associated with smaller entities due to the relationship between vocal tract length and vocal productions) \citep{sapir-1929}. Extensive research has been performed in the area of sound symbolism, demonstrating interesting findings such as these patterns being largely language agnostic \citep{cwiek-robust} as well as being weaker in neurodivergent individuals \citep{Occelli-2013} and not being yet developed in very early childhood \citep{sidhu-longitudinal}. Other research has also investigated the exact requirements and limits of the effects \citep{Sidhu2023, Passi2022, styles-fail-2017, Nielsen-consonants-2013}.

Recently, \citet{alper2023kiki} have investigated the ability of language models to exhibit the Kiki/Bouba effect using CLIP and Stable Diffusion by generating images from sound-symbolic prompts and provide positive evidence for this association to be present. We build upon this work in \S\ref{kiki-section} by introducing a wider range of VLMs in a forced naming task for ``real'' entities as opposed to abstract shapes, and additionally extend this to magnitude symbolism via the Mil-Mal test in \S\ref{sec:magnitude}. We further differentiate our work by focussing on the task of assigning pseudowords to provided visual stimuli, in contrast to \citet{alper2023kiki} who investigate the effects of different pseudowords on the outputs of image generation models.

Additionally, in recent times, large-scale efforts have been made to collect ratings of linguistic iconicity (i.e., the level of symbolism a particular word has), with \citet{Winter2023} collecting ratings of over 14k English words. Some effort has been made to analyse whether or not similar ratings would be assigned by an LLM, where \citet{trott_2023} used GPT-4 and reports a moderate positive correlation across ratings. We build upon this work in \S\ref{sec:iconicity} by introducing a wider range of VLMs, including open-source alternatives. Numerous computational works in NLP have investigated other aspects of iconicity and sound symbolism, with \citet{abramova-fernandez-2016-questioning} investigating the word embeddings of different aspects of morphology in relation to symbolism (see also \citealp{yamshchikov-etal-2019-dyr, liu-etal-2018-discovering}). Additionally, \citet{sabbatino-etal-2022-splink} investigated the emotional intensity of nonsense words using NLP methods to determine which phoneme combinations were most responsible. Several works in similar areas have also exemplified the ability of LLMs to demonstrate perceptual behaviour akin to humans and the potential for these models to replace human participants in pilot studies, as well as facilitating the scaling of evaluation \textit{in-silico} \citep{jain-2023-silico, aher2023, DILLION2023, ramezani-xu-2023-knowledge, codaforno2023inducing}.

\section{Shape Symbolism}
\label{kiki-section}
\begin{figure}[t]
\centering
\includegraphics[width=1\linewidth]{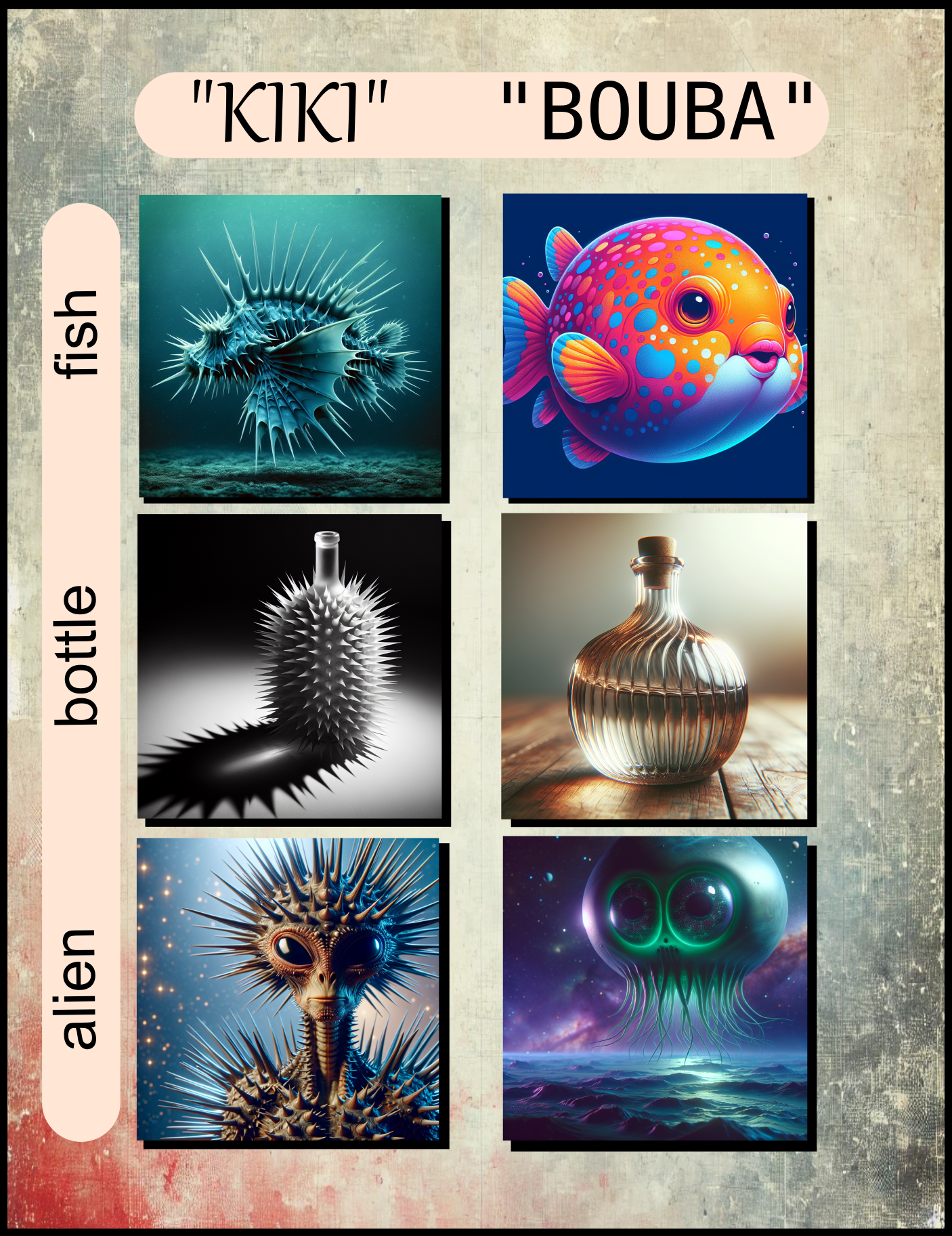}
% \includesvg[width=1.0\columnwidth]{figs/intro.svg}
\caption{Examples of ``Kiki''-style (spiky) and ``Bouba''-style (rounded) generations with DALL-E 3. In total, 50 images were generated, with 25 per condition (the entities remaining constant). The ground truth is taken as the majority human vote.}
\label{fig:kiki_image}
\end{figure}

In this section, we perform a replication of the classic Kiki-Bouba Effect experiment \cite{Ramachandran2001} using a range of multi-modal LLMs. Within the traditional set-up for the Kiki/Bouba test, human participants are presented with either a rounded soft-edged shape or a spiky sharp-edged shape and asked to assign one of two pseudowords to either. In numerous experiments \citep{Sidhu2021, Ramachandran2001,kohler-1929-takete}, words such as ``Bouba'' and ``Maluma'' are preferred for the latter rounded shapes, whilst ``Kiki'' and ``Takete'' are preferred for the spiked shapes. These findings are thought to demonstrate a non-arbitrary link between particular speech sounds and the physical characteristics of the shapes to which they refer. 

\subsection{Methodology}
\paragraph{Image Dataset}
\label{sec:kiki_dataset}
We prompt DALL-E 3 \citep{DALLE-3} to generate a series of images pertaining to entities that are either ``spiky'' or ``rounded''. Example generations can be seen in \autoref{fig:kiki_image}. We use DALL-E 3 to generate examples rather than taking existing images such as the traditional representation of Kiki-Bouba in order to reduce the effects of memorisation and exhibit finer control of the physical characteristics of the presented entities. Furthermore, this allows us to further investigate the extent to which human perception of this phenomenon extends from geometric shapes to entities in the ``real'' world, therefore increasing ecological validity and more closely representing how LLM/VLMs may be tasked with demonstrating sound symbolism in the real world (e.g., when naming new products for marketing or new characters in a narrative). We use the prompt "\textit{Generate an image from the following description: [spiky/rounded + noun]}", where we generate 25 spiky examples, and 25 round examples (each noun is used once per shape condition). The full list of entities can be seen in Appendix \ref{apx:entities}.

\paragraph{Pseudowords}
Similar to our need to generate novel imagery to better ensure the non-memorisation of our chosen VLMs, we must avoid bias from the eponymous pseudowords (i.e., ``Kiki'' or ``Takete'' for spiked concepts, or ``Bouba'' and ``Maluma'' for round concepts). As a result of this, we consult existing sound symbolism research papers and select the following pseudowords that are legal in English phonotactics and imitate the same phonetic relationship that the original terms were meant to elicit. Borrowing from \citet{Occelli-2013} we take: \textit{Kalika–Mabobe},  \textit{Zaki-Umbu}, and \textit{Tiki-Giba}. From \citet{alper2023kiki} we additionally take \textit{Kitaki-Gugagu}, \textit{Hatiha-Bodubo} and \textit{Penape-Gunogu}. Finally, we use the original \textit{Kiki-Bouba} names as a point of reference for a best-case scenario where the link is explicitly learnt from mentions within the training data.

\paragraph{Task Setting}
\label{sec:kiki-setting}
We imitate the standard human setup for the Kiki/Bouba experiment \cite{Ramachandran2001} and present our VLMs with the following zero-shot prompt -- \textit{"Look at the [ENTITY] in the provided image. Out of the following two options, which name would you most likely assign to the [ENTITY]: "[KIKI-WORD]" or "[BOUBA-WORD]". Respond with only your decision"}. One candidate from either name category (i.e., Kiki or Bouba) is presented as outlined previously, and [ENTITY] refers to a noun used to describe the entity we wish to be named in order to direct the LLM's attention to the correct element (i.e., the noun from the DALL-E prompt). For this experiment, we set $max\_tokens$ to 10 for the VLM responses and leave all other hyperparameters at default. We additionally provide an extended prompt we call \textit{informed}, which prepends "\textit{This task is related to the phenomenon of Sound Symbolism, which is a non-arbitrary relationship between the sound of a word and associations with its physical attributes}" to give the VLMs additional task knowledge. This secondary prompt scenario is used to investigate whether human-like preferences can be encouraged from the model with additional awareness of which elements of the image to focus on. We present each prompt twice, placing each pseudoword in the first or second position and then averaging the results, to mitigate positional biases in selections.

\paragraph{Models} 
We use a selection of open- and closed-source VLMs, including multimodal \textbf{GPT-4} \citep{OpenAI_GPT4_2023}, \textbf{Gemini Pro} \cite{reid2024gemini}, and \textbf{LLaVA} \cite{liu2023improved-llava}. For our open-source LLaVA model, we investigate whether sound symbolism effects arise as a direct factor of model size by including the 7-, 13-, and 34-billion parameter versions. Implementation details are given in Appendix \ref{apx:implementation}.

\paragraph{Evaluation}
As our human point of comparison, we recruited 10 human evaluators with native-level English proficiency\footnote{Evaluators were recruited in different waves following revisions. All evaluators were paid above the current UK Living Wage per hour.} via internal methods (i.e., email lists at the primary author's institution and word-of-mouth) and presented an analogous task to that which we present to the VLMs. The order of image presentation to participants is randomised to avoid order effects.

\subsection{Results}

\begin{figure*}
    \centering
    \includegraphics[width=1\linewidth]{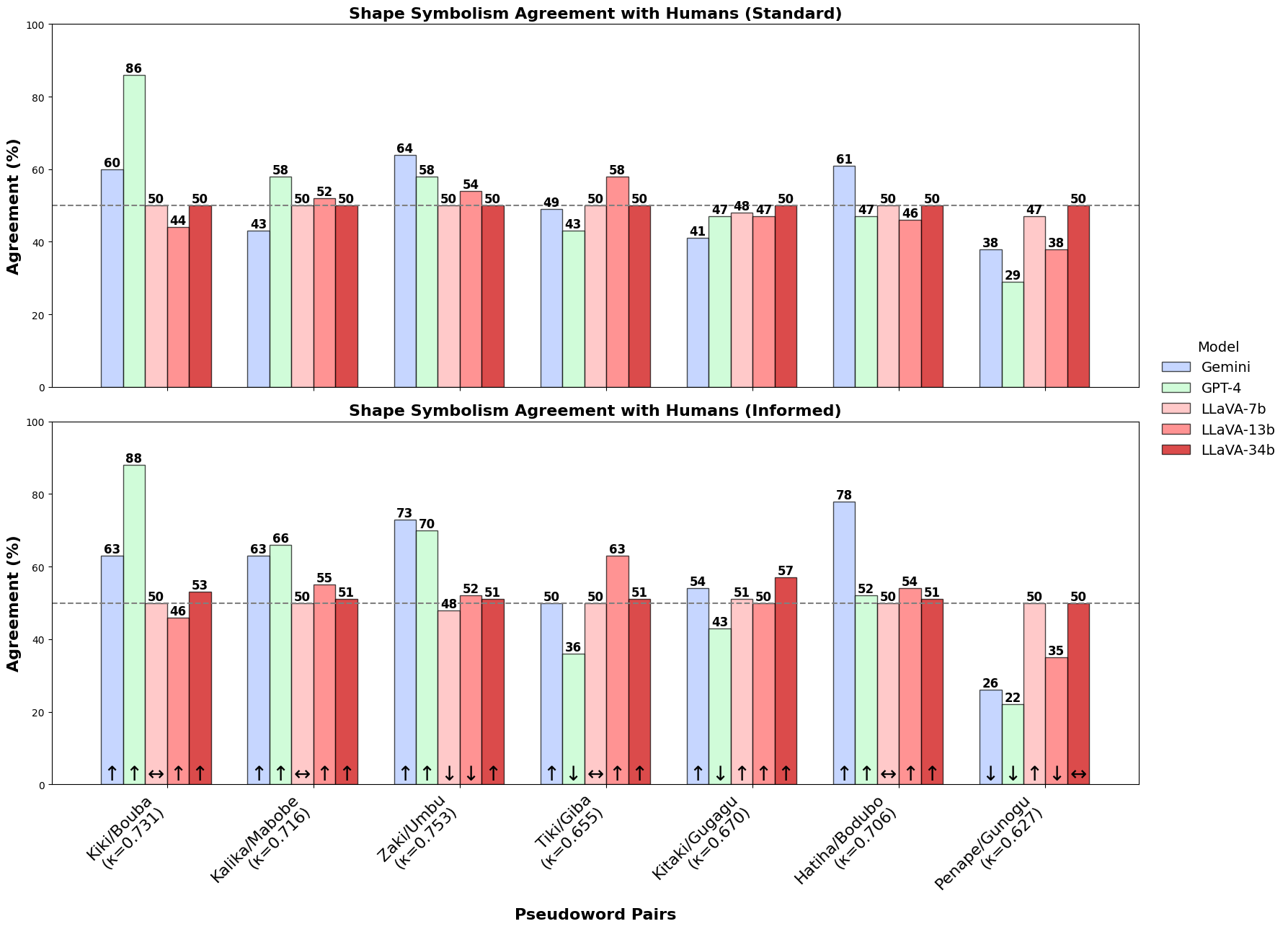}
    \caption{Results of the Shape Symbolism experiments per pseudoword pair. Fleiss' $\kappa$ \cite{fleiss_kappa} for inter-annotator agreement between humans is presented next to each pseudoword pair. Arrows indicate the direction of agreement change from the standard prompt to the informed prompt. The dashed line represents 50\%, akin to chance-level agreement. Full results in table form are presented in \autoref{tab:kiki_table} within Appendix \ref{apx:full_results}. In all cases, we are comparing with the human majority vote.}
    \label{fig:shape}
\end{figure*}

Overall, in \autoref{fig:shape} we see mixed results as to which model performs best, with GPT-4 showing the highest levels of agreement for the original Kiki-Bouba and the added Kalika-Mabobe, Gemini performing the best for Zaki-Umbu and Hatiha-Bodubo, and LLaVA performing the best for the remaining conditions. However, across all models, we see a general trend of low agreement with human ratings, with only a few condition/model combinations resulting in agreement above chance (50\%). 
Regarding the introduction of the "informed" prompt (containing additional task information), we see a general increase in agreement over the Standard condition or no change in results, indicating that once the VLMs are aware of the characteristic of interest (i.e., the shape of the entity), the VLMs are more likely to agree with human perception. However, we do see a few cases (e.g., Zaki/Umbu with LLaVA 7/13b, Kitaki/Gugagu with GPT-4, and Penape/Gugagu with Gemini, GPT-4, and LLaVA 13b) where performance decreases in the informed condition, but these are usually only minor decreases except for cases where there is already very low agreement with humans. 
We see varying performance across open and closed-source models. Whilst LLaVA outperforms GPT-4 and Gemini in many conditions, this is often close to chance-level agreement and may be the result of label bias, versus the closed models' systematic disagreement.\footnote{We see LLaVA variants demonstrate a clear preference for whatever pseudoword is presented in the first position.}
Finally, regarding our open-source LLaVA models at different sizes, we interestingly see that the 13B model outperforms the 7B and 34B models at times, most noticeably in Tiki-Giba. However, the largest 34b model performs best overall, in line with expectation (though by a small margin). Additionally, the Penape-Gunogu pair presents difficulty for our tested models, with systematic disagreement with humans by Gemini and GPT-4, suggesting the influence of additional information contained within the models' training data as to the connotations of these pseudowords.

\section{Magnitude Symbolism}
\label{sec:magnitude}

\begin{figure}[t]
\centering
\small
\includegraphics[width=1\linewidth]{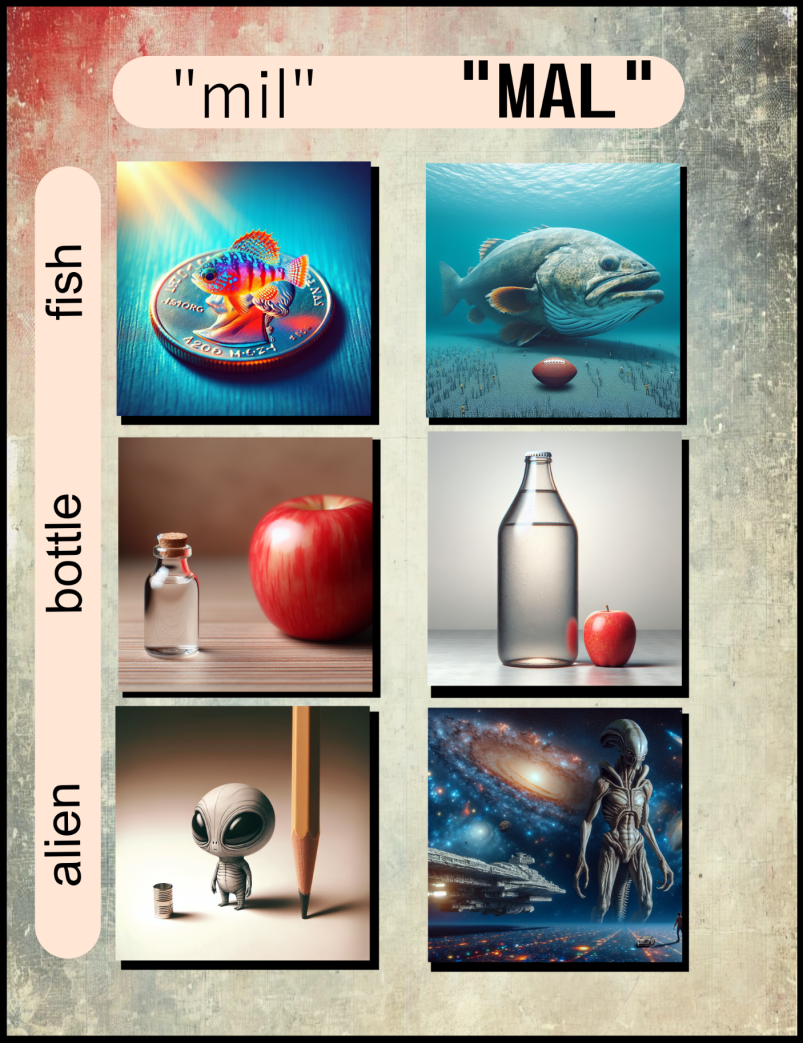}
\caption{Examples of ``Mil''-style (tiny) and ``Mal''-style (huge) generations with DALL-E 3. In total, 50 images were generated, with 25 per condition (the entities remaining constant). The ground truth is taken as the majority human vote.}
\label{fig:mil_image}
\end{figure}

Whilst the Kiki-Bouba effect demonstrates sound symbolism in relation to the perceived spikiness/roundness of an object, magnitude symbolism refers to the non-arbitrary relationship between certain vowels and the perceived physical size of the entity they refer to and is commonly demonstrated through the names ``Mil'' and ``Mal'', where the high front vowel in ``Mil'' is associated with small entities, and the low back vowel of ``Mal'' is associated with larger entities.

\subsection{Methodology}

\paragraph{Image Dataset}
We follow the same process as \S\ref{sec:kiki_dataset}, but use the characteristics of ``tiny'' and ``huge'' with the following prompt: "\textit{Generate an image of a/an [ENTITY] in isolation, with something else to help judge scale/perspective}". We use the same noun entities as in \S\ref{sec:kiki_dataset}. Example generations are in \autoref{fig:mil_image}.

\paragraph{Pseudowords}
As in the Shape Symbolism experiment (\S\ref{kiki-section}), we wish to mitigate potential bias from the memorisation capability of the VLMs. To this end, we use the ``Mil'' and ``Mal'' often associated with this test \citep{sapir-1929} in addition to other phonetically similar pseudowords. Additionally, to avoid gross extraneous factors arising from using words that are meaningful in English (such as ``mil'' referring to millilitres, and ``mal'' being associated with badness, i.e., malpractice/malnourishment), we create the minimal pairs \textit{Dil/Dal}, \textit{Zil/Zal}, \textit{Geel/Gaal}, \textit{Beel/Baal}, \textit{Weel/Waal}, and \textit{Leel/Laal}. The former three exploit the contrast between \textipa{/I/} and \textipa{/a/}, whilst the latter exploit \textipa{/i/} and \textipa{/A/}, with a range of consonants for variation.

\paragraph{Task Setting}
We use the same setup as in \S\ref{sec:kiki-setting} but present the models with one candidate from either Magnitude-based name category (i.e. ``Mil''-esque or ``Mal''-esque pseudowords), rather than Kiki/Bouba related pseudowords. We additionally provide an extended prompt we call \textit{informed}, which prepends "\textit{This task is related to the phenomenon of Magnitude Symbolism, which is a non-arbitrary relationship between the sound of a word and its association with size and scale}" to give additional task knowledge.

\paragraph{Models \& Evaluation}
We use the same models and evaluation protocols as in \S\ref{sec:kiki-setting}.

\subsection{Results}

\begin{figure*}
    \centering
    \includegraphics[width=1\linewidth]{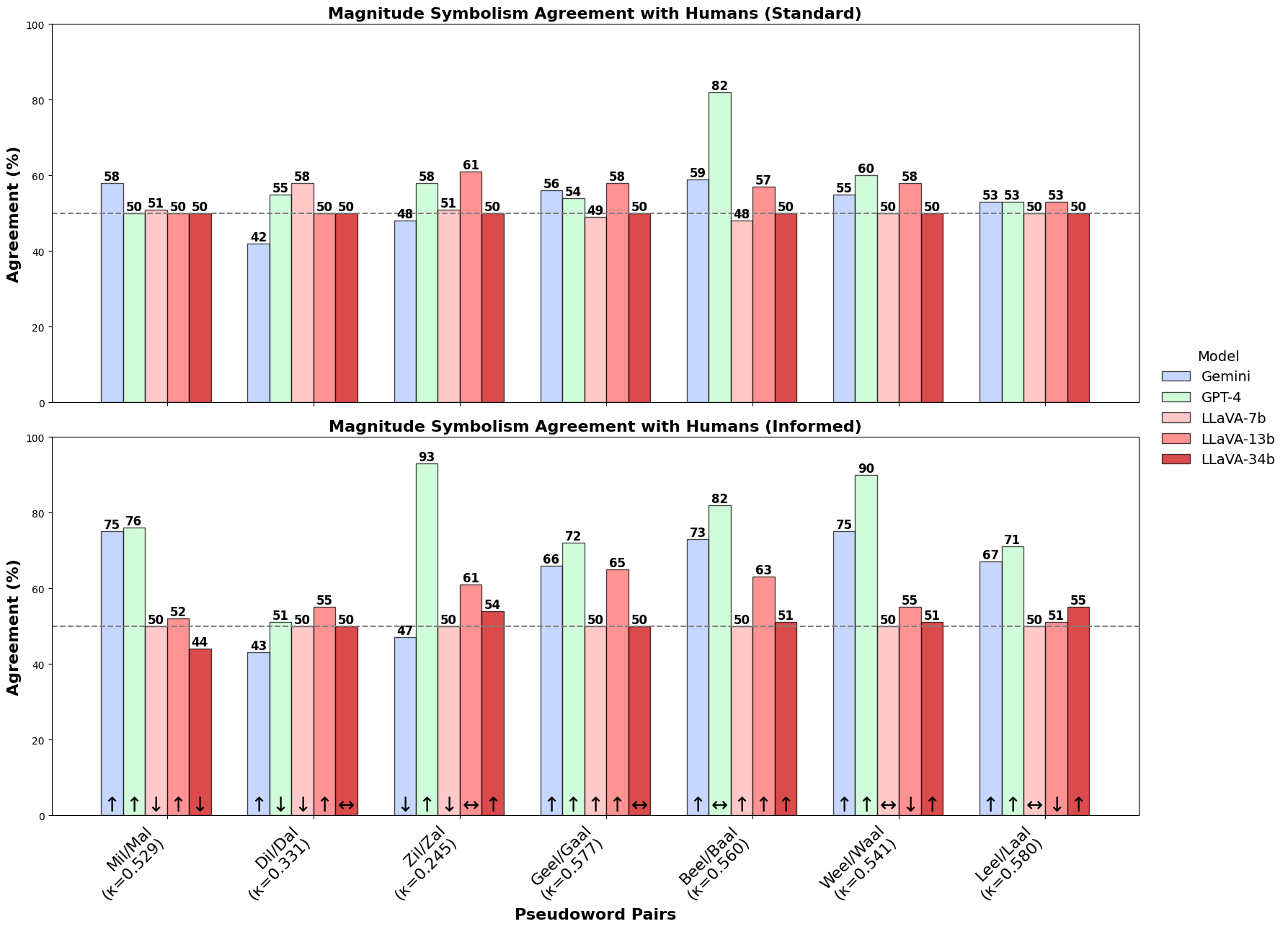}
    \caption{Results of the Magnitude Symbolism experiments per pseudoword pair. Fleiss' $\kappa$ \cite{fleiss_kappa} for inter-annotator agreement between humans is presented next to each pseudoword pair. Arrows indicate the direction of agreement change from the standard prompt to the informed prompt. The dashed line represents 50\%, akin to chance-level agreement. Full results in table form are presented in \autoref{tab:mil_table} within Appendix \ref{apx:full_results}. In all cases, we are comparing with the human majority vote.}
    \label{fig:magnitude}
\end{figure*}

Overall, similar to \S\ref{kiki-section} we see mixed results in \autoref{fig:magnitude} as to which model performs best, but GPT-4 demonstrates higher levels of agreement across most conditions in a clearer pattern than what was seen in the Shape Symbolism experiment. In several cases, we see agreement that is significantly in line with human perception, with 90+\% agreement (e.g., GPT-4 in the Zil-Zal and Weel-Waal conditions). When comparing the \textit{standard} and \textit{informed} prompts, we see a much more substantial increase in performance, indicating that the VLMs fundamentally understand the relationship between sound and perceived size, but were focused on other aspects of the provided imagery when not explicitly directed towards size in the \textit{standard} condition. Regarding the LLaVA models, we see the mid-sized 13B variant outperforming the 7B and 34B models in most conditions (rather than performance increasing alongside parameter count).

\section{Iconicity Ratings}
\label{sec:iconicity}
In this section we investigate whether LLMs demonstrate human-like associations between word forms and the entities/concepts they symbolise.\footnote{We treat this as VLMs ``imitating'' an understanding of sound symbolism, as they of course cannot actually hear.} \citet{Winter2023} present a dataset of 14k+ human judgements of iconicity to which we compare our LLM ratings (where, on a 7-point scale, ``how'' scored 1.3, whilst ``woof'' scored 6.8 due to being onomatopoeic). 

\subsection{Methodology}
\paragraph{Models} 
We use a range of modern LLMs for this task. Specifically, \textbf{GPT-4} \citep{OpenAI_GPT4_2023}, \textbf{GPT-3.5-Turbo} \citep{RLHF-openai}, \textbf{LLaMA-2} (7B, 13B and 70B) \citep{LLAMA-2}, \textbf{FLAN-T5} (base and XL) \citep{T5,FLAN}, and \textbf{Mistral-7B} \citep{jiang2023mistral}. Implementation details are presented in Appendix \ref{apx:implementation}.

Existing work by \citet{trott_2023} has investigated whether or not GPT-4 is able to reflect human judgements and reported a Spearman correlation of 0.59. However, in trying to verify these findings, we see that our ratings differ quite strongly. For this reason, we re-run this experiment on the first 7k entries of \citet{Winter2023} and compare GPT-4 ratings as of late November 2023 with the human judgements in \citet{Winter2023} as well as the GPT-4 judgements from \citet{trott_2023} that use an earlier version of GPT-4.\footnote{We test only on the first 7k entries due to cost. Whilst GPT-4-Turbo is markedly cheaper, this would not be a true reproduction. Please note that GPT-4o was not released at the time.} In the other cases, we use the entire 14,772 entry dataset, except for GPT-3.5-Turbo, where we remove 209 words that triggered OpenAI's safeguarding filters.

\paragraph{Prompting}
We adopt the same prompting strategy as \citet{trott_2023}. In doing so, we use a modified version of what was presented to human participants in \citet{Winter2023}. In summary, we request ratings of iconicity on a 1-7 scale, where 1 is not at all iconic, and 7 is highly iconic. The full prompt presented to the models is available in Appendix \ref{apx:prompt}.

\subsection{Results}
Correlations between each LLM and human judgements are presented in \autoref{tab:iconicity}. We observe that the ability of LLMs to rate the iconicity of English words appears to be dependent on model size. For instance, we see no true correlation with any of our FLAN T5 models (base = 250M, XL = 3B) or our smallest LLaMA-2 variant (7B), but we begin to see significant correlations with another 7B parameter model, that being instruction-tuned Mistral-7B. When investigating our larger models, we see the 13B LLaMA-2 variant demonstrate Spearman/Pearson correlations of .379 and .381, respectively. Interestingly, however, our largest LLaMA-2 model with 70B parameters performs worse at this task than the 13B variant as seen previously, with correlations of .304/.332. Regarding OpenAI models, GPT-3.5-Turbo demonstrates a moderate correlation with humans at .420/.439. Finally, GPT-4  presents the strongest correlations. However, we observe that (on the first 7k entries), the version of GPT-4 we use (late November 2023) performs worse than the earlier version presented by \citet{trott_2023}, further demonstrating how continuously updated models also require continuous evaluation due to receiving additional training data (Spearman correlations of 0.537 vs 0.575, respectively).

\begin{table}[h]
\small
\centering
\begin{tabular}{lcccc}
\toprule
& \multicolumn{2}{c}{\textbf{Spearman}} & \multicolumn{2}{c}{\textbf{Pearson}} \\ \cmidrule{2-5}
\textbf{Model} & Corr. & $p$ & Corr. & \(p\) \\ \midrule
FLAN-T5 Base & -.035 & \textless{}.001 & -.037 & \textless{}.001 \\ 
FLAN-T5 XL & .000 & .991 & -.002 & .824 \\ 
Mistral-7B-Instruct & .382 & \textless{}.001 & .377 & \textless{}.001 \\
LLaMA-2 7B & -.003 & .687 & .005 & .544 \\ 
LLaMA-2 13B& .379 & \textless{}.001 & .381 & \textless{}.001\\ 
LLaMA-2 70B & .304 & \textless{}.001 & .332 & \textless{}.001 \\ 
GPT-3.5-Turbo & .420 & \textless{}.001 & .439 & \textless{}.001 \\ 
GPT-4 \citep{trott_2023} & \textbf{.575} & \textless{}.001 & \textbf{.615} & \textless{}.001\\
GPT-4 (Ours, Nov. '23) & \textbf{.537} & \textless{}.001 & \textbf{.594} & \textless{}.001\\ \bottomrule
\end{tabular}
\caption{Correlations between human ratings from \citet{Winter2023} and LLMs. GPT-4 ratings were only collected for the first 7k examples due to cost. \citet{trott_2023} report a Spearman correlation of .590 across the entire dataset. \textit{p} refers to p-value.}
\label{tab:iconicity}
\end{table}

\section{Discussion}
\label{sec:discussion}
\paragraph{The Source of Sound Symbolism} Across our experiments, we see evidence that LLM/VLMs are capable of making decisions that are similar to those of humans in sound symbolism tasks, whilst only having access to textual and visual modalities, while human decisions are believed to be grounded in sound. We hypothesise several reasons for the emergence of sound symbolism in LLM/VLMs.

Firstly, due to human languages exhibiting mostly regular orthography, auditory information in speech is moderately reflected in the spellings of words via grapheme sequences (a characteristic that grapheme-to-phoneme conversion models have long exploited). Through this, text-based models are able to learn associations between grapheme sequences and semantics, based on more abstract characteristics than morpheme combinations alone, such as phonaesthemes \cite{kaushal-mahowald-2022-tokens}. Whilst such models have no embodied understanding of sound, such statistical patterns pose a viable signal for the implicit learning of sound-based phenomena. 

Secondly, such associations between sounds (or grapheme combinations) and physical characteristics are naturally present in language, such as in poetry, narratives, or descriptions of entities that are cute, scary, small, or large, and are consequently paired with relevant visual stimuli in image captions when training vision modules for multimodal systems. However, such associations are subtle and not entirely ubiquitous. For example, whilst the high front vowel \textipa{/I/} typically associated with small entities is present in "tiny" and "mini", the word "small" itself possesses a low back vowel \textipa{/O:/}. 

As a result, the relatively weak performance of our tested models could also be explained by the relative lack of sound-symbolism-heavy language in the models' training data which is overshadowed by more prosaic language forms that do not exploit these phenomena as readily. This in turn would explain why the closed-source models we tested (e.g. GPT-4/Gemini) outperform open-source models due to the significant (assumed) differences in parameter size, allowing the larger closed models to retain more information regarding sound symbolism within the weights, in addition to being continuously updated with RLHF. 

The results of our multimodal experiments additionally demonstrate that under certain conditions, VLMs show systematic disagreement with human labels, indicating the potential interference of additional knowledge contained within language model training data that influences the associations made between pseudowords and images that are not present in humans. However, it is important to note that in our experiments we compare language model selections against the majority vote or mean scores assigned by humans. Consequently, this results in a comparison to an "ideal" human by necessity, overlooking individual differences in perception across humans (where for a decision to be "human-like", it has to match a choice made by any human, rather than the majority). Consequently, higher agreement levels can be observed when compared to the choices of individual humans, as inter-human agreement is not perfectly aligned in these tests.

\paragraph{Future Directions} As a result of LLM/VLMs not being able to fully reflect human preferences in tasks regarding sound symbolism, it remains a promising future direction to explicitly pre-train language models on more sound-symbolism heavy datasets or explicitly include sound-symbolism-related tasks into the training or finetuning of these models for use on related downstream tasks (such as creative writing and marketing). Additionally, investigating the reason behind model predictions is a promising direction, such as through additional prompting to generate textual justifications, or investigating the visual attention of VLMs to investigate whether they are attending to characteristics closely associated with the concepts being tested (e.g., spikes).

\section{Conclusion}
We have shed light on the processes underlying multimodal perception and understanding in language models. To do so, we performed a series of tests on modern VLM/LLMs regarding their ability to exhibit an understanding of sound symbolism. Through comparison with human judgements, we see that VLMs are able to approximate human perception in sound symbolism tests under certain conditions, such as when informed of the nature of the study (via the \textit{informed} prompts), but struggle overall. We additionally see that magnitude symbolism potentially presents an easier pattern for VLMs to recognise than shape symbolism, with selections having a higher agreement with humans on Magnitude Symbolism tests than Shape Symbolism. We also see that the ability of LLMs to emulate human judgements of iconicity scales more linearly with model size. These findings indicate room for future research on more explicit inclusion of abstract perceptual properties into language model training in order to facilitate better \textit{in silico} experimentation and improve performance on other downstream tasks.

\section*{Limitations}
Owing to the relatively small sample sizes (i.e., the number of pseudoword pairs in the VLM-related tasks), we treat this work as a proof-of-concept as to the ability of LLMs to perform well in the tasks we present and encourage other parties to engage in similar research at scale if their situation permits. Additionally, whilst sound symbolism is believed to be largely language agnostic, we only use native English speakers and pseudowords that are phonotactically legal in English in the present work. Additionally, some of our chosen pseudowords are taken from existing literature. Whilst we investigated the prevalence of these words in the context of sound symbolism within internet resources in order to mitigate memorisation from training data, it remains possible that some level of data contamination may be present (although the overall low performance casts doubt on this). Furthermore, we present only the orthographic forms of the pseudowords to human participants, resulting in potential variation between speakers regarding phonetic realisation.

\section*{Ethics Statement}
We believe in and firmly adhere to the Code of Conduct in the performance of this work and the methods involved. All of our imagery generations were provided via accessing the respective OpenAI APIs, and in discovering imagery that triggered OpenAI's built-in guardrails, we replaced these images with other options. All human evaluation was performed by consenting adult participants who were provided with a participant information sheet and subsequently signed a consent form in line with the Ethics procedures of the primary author's institution (who approved the ethical validity of the study performed herein). Additionally, we present this work as a demonstration of interesting behaviours within (very) large LLMs, but do not condone the wholesale replacement of human participants in related psycholinguistic/cognitive/psychological experimentation, but rather view \textit{in silico} experimentation as a useful tool primarily for prototyping.

\subsubsection*{Acknowledgments}
Tyler Loakman is supported by the Centre for Doctoral Training in Speech and Language Technologies (SLT) and their Applications funded by UK Research and Innovation [grant number EP/S023062/1].

Visual elements in \autoref{fig:first-page} were taken from Flaticon. Specifically, banners (SANB), brain (Freepik), GPU (Taufik Ramadhan), "versus" (Afif Fudin), descending/ascending (Mie Nakae), and thought bubble (Aranagraphics).

\bibliography{custom}

\appendix
\section{Appendix}

\subsection{Implementation Details}
\label{apx:implementation}

\paragraph{Kiki-Bouba \& Mil-Mal}

For the Kiki-Bouba and Mil-Mal sound symbolism experiments, we access \textit{gemini-pro-vision} via the Google Gemini API. For GPT-4, we use \textit{gpt-4-vision-preview} via the OpenAI Chat Completions API. For our open-source LLaVA models at various sizes, we specifically use \textit{llava-v1.6-vicuna-7b-hf}, \textit{LLaVA-v1.6-vicuna-13b-hf} and \textit{LLaVA-v1.6-vicuna-34b-hf} from publicly available checkpoints on Hugging Face. We use default hyperparameters for all models to test their ``off-the-shelf'' capability. Human participants were shown generated imagery with the binary pseudo-word options via Google Forms. The order of image presentation was randomised, whilst the order of the pseudowords was kept static. Separate Google Forms were used for each pseudoword pair. Participants were able to complete the forms at their own pace within a period of approximately 2 weeks.

\paragraph{Iconicity Ratings}

For our iconicity rating experiments, we use the following models from Hugging Face: FLAN-T5 Base (\textit{google/flan-t5-base}), FLAN-T5 XL (\textit{google/flan-t5-xl}) Mistral-7B (\textit{mistralai/Mistral-7B-Instruct-v0.2}), LLaMA-2 (\textit{meta-llama/Llama-2-7b-hf, meta-llama/Llama-2-13b-hf, meta-llama/Llama-2-70b-hf}). We access GPT-3.5-Turbo and GPT-4 via the OpenAI Chat Completions API. We also keep all model hyperparameters at default settings. For the LLaVA models, we modify the prompt slightly by adding choice labels (A/B) rather than requesting the pseudoword itself to be returned in order to directly access single-token output probabilities.

\paragraph{Entities}
\label{apx:entities}
Within our DALL-E 3 generations in the aforementioned experiments, we select the following list of entities in order to have a range of characteristics (including animate and inanimate entities): \textit{alien, bed, bird, bottle, cat, chair, desk, dog, door, fish, flower, fruit, ghost, house, insect, lizard, machine, person, plane, planet, snake, toy, tree, vegetable,} and \textit{vehicle}.

\subsection{Full Results}
\label{apx:full_results}
\begin{table*}[htbp]
\centering
\small
\begin{tabular}{lccccc}
\toprule
\textbf{Prompt} & \multicolumn{5}{c}{\textbf{Kiki/Bouba ($\kappa$ .731)}}\\
\midrule
 & \textbf{Gemini $\uparrow$} & \textbf{GPT-4 $\uparrow$} & \textbf{LLaVA-7b} & \textbf{LLaVA-13b $\uparrow$} & \textbf{LLaVA-34b $\uparrow$} \\
\textbf{Standard} & 60\% ($\kappa$ .200) &\textbf{86\% ($\kappa$ .720)} & \underline{50\% ($\kappa$ .000)} & 44\% ($\kappa$ -.120) & \underline{50\% ($\kappa$ .000)} \\
\textbf{Informed} & 63\% ($\kappa$ .260) & \textbf{88\% ($\kappa$ .760)} & 50\% ($\kappa$ .000) & 46\% ($\kappa$ -.080) & \underline{53\% ($\kappa$ .060)} \\
\midrule
\textbf{Prompt} & \multicolumn{5}{c}{\textbf{Kalika/Mabobe ($\kappa$ .716)}}\\
\midrule
 & \textbf{Gemini $\uparrow$} & \textbf{GPT-4 $\uparrow$} & \textbf{LLaVA-7b} & \textbf{LLaVA-13b $\uparrow$} & \textbf{LLaVA-34b $\uparrow$} \\
\textbf{Standard} & 43\% ($\kappa$ -.140) & \textbf{58\% ($\kappa$ .160)} & 50\% ($\kappa$ .000) & \underline{52\% ($\kappa$ .040)} & 50\% ($\kappa$ .000) \\
\textbf{Informed} & 63\% ($\kappa$ .260) & \textbf{66\% ($\kappa$ .320)} & 50\% ($\kappa$ .000) & \underline{55\% ($\kappa$ .100)} & 51\% ($\kappa$ .020) \\
\midrule
\textbf{Prompt} & \multicolumn{5}{c}{\textbf{Zaki/Umbu ($\kappa$ .753)}}\\
\midrule
 & \textbf{Gemini $\uparrow$} & \textbf{GPT-4 $\uparrow$} & \textbf{LLaVA-7b $\downarrow$} & \textbf{LLaVA-13b $\downarrow$} & \textbf{LLaVA-34b $\uparrow$} \\
\textbf{Standard} & \textbf{64\% ($\kappa$ .277)} & 58\% ($\kappa$ .152) & 50\% ($\kappa$ .002) & \underline{54\% ($\kappa$ .103)} & 50\% ($\kappa$ .000) \\
\textbf{Informed} & \textbf{73\% ($\kappa$ .464)} & 70\% ($\kappa$ .402) & 48\% ($\kappa$ .000) & \underline{52\% ($\kappa$ .065)} & 51\% ($\kappa$ .019) \\
\midrule
\textbf{Prompt} & \multicolumn{5}{c}{\textbf{Tiki/Giba ($\kappa$ .655)}}\\
\midrule
 & \textbf{Gemini $\downarrow$} & \textbf{GPT-4 $\downarrow$} & \textbf{LLaVA-7b} & \textbf{LLaVA-13b $\uparrow$} & \textbf{LLaVA-34b $\uparrow$} \\
\textbf{Standard} & 49\% ($\kappa$ -.020) & 43\% ($\kappa$ -.140) & 50\% ($\kappa$ .000) & \underline{\textbf{58\% ($\kappa$ .160)}} & 50\% ($\kappa$ .000) \\
\textbf{Informed} & 50\% ($\kappa$ .000) & 36\% ($\kappa$ -.280) & 50\% ($\kappa$ .000) & \underline{\textbf{63\% ($\kappa$ .260)}} & 51\% ($\kappa$ .020) \\
\midrule
\textbf{Prompt} & \multicolumn{5}{c}{\textbf{Kitaki/Gugagu ($\kappa$ .670)}}\\
\midrule
 & \textbf{Gemini $\uparrow$} & \textbf{GPT-4 $\downarrow$} & \textbf{LLaVA-7b $\uparrow$} & \textbf{LLaVA-13b $\uparrow$} & \textbf{LLaVA-34b $\uparrow$} \\
\textbf{Standard} & 41\% ($\kappa$ -.180) & 47\% ($\kappa$ -.060) & 48\% ($\kappa$ -.040) & 47\% ($\kappa$ -.060) & \underline{\textbf{50\% ($\kappa$ .000)}} \\
\textbf{Informed} & 54\% ($\kappa$ .080) & 43\% ($\kappa$ -.140) & 51\% ($\kappa$ .020) & 50\% ($\kappa$ .000) & \underline{\textbf{57\% ($\kappa$ .140)}} \\
\midrule
\textbf{Prompt} & \multicolumn{5}{c}{\textbf{Hatiha/Bodubo ($\kappa$ .706)}}\\
\midrule
 & \textbf{Gemini $\uparrow$} & \textbf{GPT-4 $\uparrow$} & \textbf{LLaVA-7b} & \textbf{LLaVA-13b $\uparrow$} & \textbf{LLaVA-34b $\uparrow$} \\
\textbf{Standard} & \textbf{61\% ($\kappa$ .220)} & 47\% ($\kappa$ -.060) & \underline{50\% ($\kappa$ .000)} & 46\% ($\kappa$ -.080) & \underline{50\% ($\kappa$ .000)} \\
\textbf{Informed} & \textbf{78\% ($\kappa$ .560)} & 52\% ($\kappa$ .040) & 50\% ($\kappa$ .000) & \underline{54\% ($\kappa$ .080)} & 51\% ($\kappa$ .020) \\
\midrule
\textbf{Prompt} & \multicolumn{5}{c}{\textbf{Penape/Gunogu ($\kappa$ .627)}}\\
\midrule
 & \textbf{Gemini $\downarrow$} & \textbf{GPT-4 $\downarrow$} & \textbf{LLaVA-7b $\uparrow$} & \textbf{LLaVA-13b $\downarrow$} & \textbf{LLaVA-34b} \\
\textbf{Standard} & 38\% ($\kappa$ -.240) & 29\% ($\kappa$ -.420) & 47\% ($\kappa$ -.060) & 38\% ($\kappa$ -.240) & \underline{\textbf{50\% ($\kappa$ .000)}} \\
\textbf{Informed} & 26\% ($\kappa$ -.480) & 22\% ($\kappa$ -.560) & 50\% ($\kappa$ .000) & 35\% ($\kappa$ -.300) & \underline{\textbf{50\% ($\kappa$ .000)}} \\
\midrule
\textbf{Prompt} & \multicolumn{5}{c}{\textbf{ALL (excl. Kiki/Bouba)}}\\
\midrule
 & \textbf{Gemini $\uparrow$} & \textbf{GPT-4 $\uparrow$} & \textbf{LLaVA-7b $\downarrow$} & \textbf{LLaVA-13b $\uparrow$} & \textbf{LLaVA-34b $\uparrow$} \\
\textbf{Standard} & 49.33\% ($\kappa$ -.014) & 47.83\% ($\kappa$ -.045) & 49.17\% ($\kappa$ -.016) & 49.71\% ($\kappa$ -.013) & \underline{\textbf{50.40\% ($\kappa$ .012)}} \\
\textbf{Informed} & \textbf{57.33\% ($\kappa$ .147)} & 48.17\% ($\kappa$ -.036) & 50.17\% ($\kappa$ .003) & 51.50\% ($\kappa$ .034) & \underline{51.83\% ($\kappa$ .037)} \\
\bottomrule
\end{tabular}
\caption{
Results of the Shape Symbolism experiments per pseudoword pair. Fleiss' $\kappa$ \cite{fleiss_kappa} for inter-annotator agreement between humans is presented next to each pseudoword pair. For each VLM and word pair, we present Cohen's $\kappa$ for agreement between the models and the human majority vote \cite{cohens_kappa}. The model with the highest agreement per prompt is in \textbf{bold}, and the best performing open-source model (i.e., LLaVA variant) is \underline{underlined}. Arrows next to model names indicate the direction of agreement change from the standard prompt to the informed prompt.}
\label{tab:kiki_table}
\end{table*}

\begin{table*}[htbp]
\centering
\small
\begin{tabular}{lccccc}
\toprule
\textbf{Prompt} & \multicolumn{5}{c}{\textbf{Mil/Mal ($\kappa$ .529)}}\\
\midrule
 & \textbf{Gemini $\uparrow$} & \textbf{GPT-4 $\uparrow$} & \textbf{LLaVA-7b $\downarrow$} & \textbf{LLaVA-13b $\uparrow$} & \textbf{LLaVA-34b $\downarrow$} \\
\textbf{Standard} & \textbf{58\% ($\kappa$ .152)} & 50\% ($\kappa$ .031) & \underline{51\% ($\kappa$ .018)} & 50\% ($\kappa$ .000) & 50\% ($\kappa$ .000) \\
\textbf{Informed} & 75\% ($\kappa$ .512) & \textbf{76\% ($\kappa$ .529)} & 50\% ($\kappa$ .000) & \underline{52\% ($\kappa$ .043)} & 44\% ($\kappa$ -.109) \\
\midrule
\textbf{Prompt} & \multicolumn{5}{c}{\textbf{Dil/Dal ($\kappa$ .331)}}\\
\midrule
 & \textbf{Gemini $\downarrow$} & \textbf{GPT-4 $\downarrow$} & \textbf{LLaVA-7b $\downarrow$} & \textbf{LLaVA-13b $\uparrow$} & \textbf{LLaVA-34b} \\
\textbf{Standard} & 42\% ($\kappa$ -.168) & 55\% ($\kappa$ .098) & \textbf{\underline{58\% ($\kappa$ .149)}} & 50\% ($\kappa$ .000) & 50\% ($\kappa$ .000) \\
\textbf{Informed} & 43\% ($\kappa$ -.128) & 51\% ($\kappa$ .010) & 50\% ($\kappa$ .000) & \underline{\textbf{55\% ($\kappa$ .122)}} & 50\% ($\kappa$ .000) \\
\midrule
\textbf{Prompt} & \multicolumn{5}{c}{\textbf{Zil/Zal ($\kappa$ .245)}}\\
\midrule
 & \textbf{Gemini $\downarrow$} & \textbf{GPT-4 $\uparrow$} & \textbf{LLaVA-7b $\downarrow$} & \textbf{LLaVA-13b} & \textbf{LLaVA-34b $\uparrow$} \\
\textbf{Standard} & 48\% ($\kappa$ -.003) & 58\% ($\kappa$ .149) & 51\% ($\kappa$ .022) & \underline{\textbf{61\% ($\kappa$ .222)}} & 50\% ($\kappa$ .000) \\
\textbf{Informed} & 47\% ($\kappa$ -.031) & \textbf{93\% ($\kappa$ .860)} & 50\% ($\kappa$ .000) & \underline{61\% ($\kappa$ .207)} & 54\% ($\kappa$ .083) \\
\midrule
\textbf{Prompt} & \multicolumn{5}{c}{\textbf{Geel/Gaal ($\kappa$ .577)}}\\
\midrule
 & \textbf{Gemini $\uparrow$} & \textbf{GPT-4 $\uparrow$} & \textbf{LLaVA-7b $\downarrow$} & \textbf{LLaVA-13b $\uparrow$} & \textbf{LLaVA-34b} \\
\textbf{Standard} & 56\% ($\kappa$ .120) & 54\% ($\kappa$ .080) & 49\% ($\kappa$ -.020) & \textbf{\underline{58\% ($\kappa$ .160)}} & 50\% ($\kappa$ .000) \\
\textbf{Informed} & 66\% ($\kappa$ .320) & \textbf{72\% ($\kappa$ .440)} & 50\% ($\kappa$ .000) & \underline{65\% ($\kappa$ .300)} & 50\% ($\kappa$ .000) \\
\midrule
\textbf{Prompt} & \multicolumn{5}{c}{\textbf{Beel/Baal ($\kappa$ .560)}}\\
\midrule
 & \textbf{Gemini $\uparrow$} & \textbf{GPT-4} & \textbf{LLaVA-7b} & \textbf{LLaVA-13b $\downarrow$} & \textbf{LLaVA-34b} \\
\textbf{Standard} & 59\% ($\kappa$ .180) & \textbf{82\% ($\kappa$ .640)} & 48\% ($\kappa$ -.040) & \underline{57\% ($\kappa$ .140)} & 50\% ($\kappa$ .000) \\
\textbf{Informed} & 73\% ($\kappa$ .460) & \textbf{82\% ($\kappa$ .640)} & 50\% ($\kappa$ .000) & \underline{63\% ($\kappa$ .260)} & 51\% ($\kappa$ .020) \\
\midrule
\textbf{Prompt} & \multicolumn{5}{c}{\textbf{Weel/Waal ($\kappa$ .541)}}\\
\midrule
 & \textbf{Gemini $\uparrow$} & \textbf{GPT-4 $\uparrow$} & \textbf{LLaVA-7b} & \textbf{LLaVA-13b $\downarrow$} & \textbf{LLaVA-34b $\uparrow$} \\
\textbf{Standard} & 55\% ($\kappa$ .100) & \textbf{60\% ($\kappa$ .200)} & 50\% ($\kappa$ .000) & \underline{58\% ($\kappa$ .160)} & 50\% ($\kappa$ .000) \\
\textbf{Informed} & 75\% ($\kappa$ .500) & \textbf{90\% ($\kappa$ .800)} & 50\% ($\kappa$ .000) & \underline{55\% ($\kappa$ .100)} & 51\% ($\kappa$ .020) \\
\midrule
\textbf{Prompt} & \multicolumn{5}{c}{\textbf{Leel/Laal ($\kappa$ .580)}}\\
\midrule
 & \textbf{Gemini $\uparrow$} & \textbf{GPT-4 $\uparrow$} & \textbf{LLaVA-7b} & \textbf{LLaVA-13b $\downarrow$} & \textbf{LLaVA-34b $\uparrow$} \\
\textbf{Standard} & \textbf{53\% ($\kappa$ .060)} &\textbf{53\% ($\kappa$ .060)} & 50\% ($\kappa$ .000) & \textbf{\underline{53\% ($\kappa$ .060)}} & 50\% ($\kappa$ .000) \\
\textbf{Informed} & 67\% ($\kappa$ .340) & \textbf{71\% ($\kappa$ .420)} & 50\% ($\kappa$ .000) & 51\% ($\kappa$ .020) & \underline{55\% ($\kappa$ .100)} \\
\midrule
\textbf{Prompt} & \multicolumn{5}{c}{\textbf{ALL (excl. Mil-Mal)}}\\
\midrule
 & \textbf{Gemini $\uparrow$} & \textbf{GPT-4 $\uparrow$} & \textbf{LLaVA-7b $\downarrow$} & \textbf{LLaVA-13b $\uparrow$} & \textbf{LLaVA-34b $\uparrow$} \\
\textbf{Standard} & 52.17\% ($\kappa$ .048) & \textbf{58.50\% ($\kappa$ .169)} & 51.00\% ($\kappa$ .019) & \underline{56.17\% ($\kappa$ .127)} & 50.00\% ($\kappa$ .000) \\
\textbf{Informed} & 61.83\% ($\kappa$ .243) & \textbf{76.50\% ($\kappa$ .528)} & 50.00\% ($\kappa$ .000) & \underline{58.33\% ($\kappa$ .168)} & 51.83\% ($\kappa$ .037) \\
\bottomrule
\end{tabular}
\caption{
Results of the Magnitude Symbolism experiments per pseudoword pair. Fleiss' $\kappa$ \cite{fleiss_kappa} for inter-annotator agreement between humans is presented next to each pseudoword pair. For each VLM, we present Cohen's $\kappa$ for agreement between the models and the human majority vote \cite{cohens_kappa}. The model with the highest agreement per prompt is in \textbf{bold}, and the best performing open-source model (i.e., LLaVA variant) is \underline{underlined}. Arrows next to model names indicate the direction of agreement change from the standard prompt to the informed prompt.
}
\label{tab:mil_table}
\end{table*}

\paragraph{Shape Symbolism (Kiki/Bouba)} \autoref{tab:kiki_table} presents the full unabridged results for the Sound Symbolism experiments presented in \S \ref{kiki-section}.

\paragraph{Magnitude Symbolism (Mil/Mal)} \autoref{tab:mil_table} presents the full unabridged results for the Magnitude Symbolism experiments presented in \S \ref{sec:magnitude}.

\section{Soundscape Description}
\label{apx:soundscape}

\begin{figure}
\small
\centering
\includegraphics[width=1\linewidth]{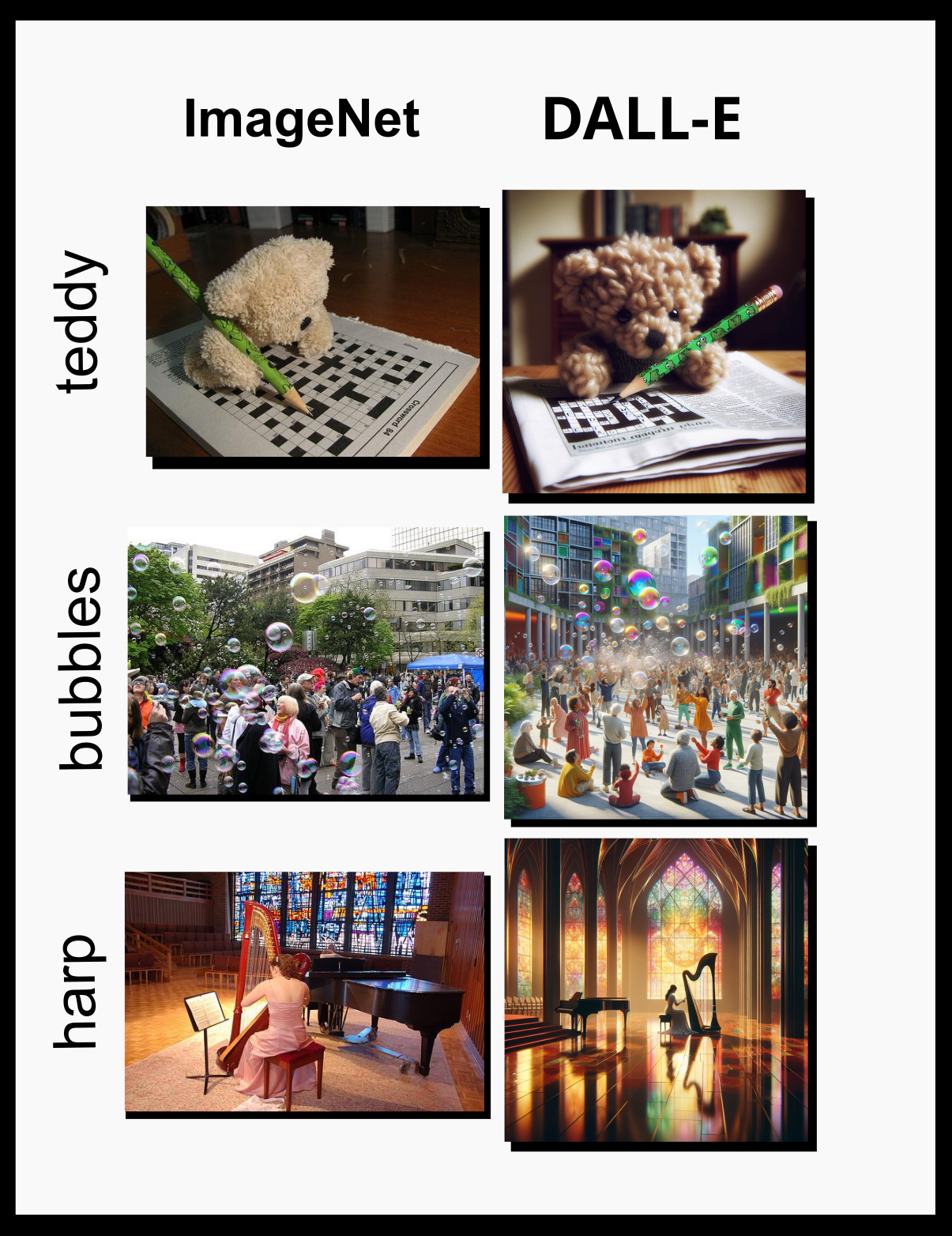}
\caption{Examples of ImageNet and DALL-E 3 image pairings for both the soundscape description and backtranslation experiments. Here, the GPT-4 description of the ImageNet image has been used in a prompt to generate the novel DALL-E 3 image.}
\label{fig:soundscape_image}
\end{figure}

\begin{figure*}[t]
\centering
\small
\includegraphics[width=0.75\linewidth]{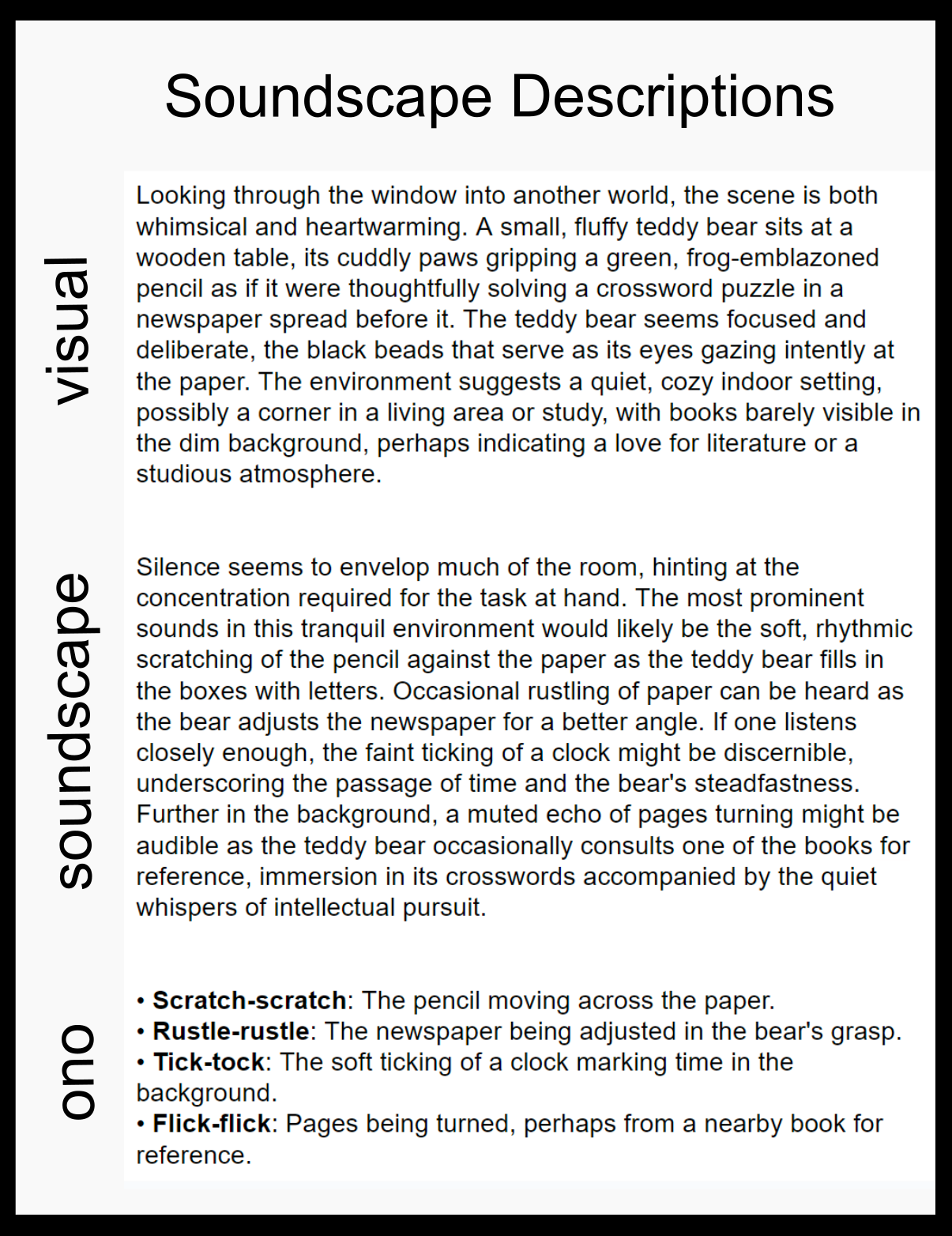}
\caption{An example output from GPT-4 when asked to describe the visuals, soundscape, and perceived onomatopoeia of an image. In this case, the image is the ImageNet generation of a teddy solving a crossword seen in \autoref{fig:soundscape_image}. \textit{ono} refers to onomatopoeia.}
\label{fig:soundscape_example}
\end{figure*}

We additionally ask to what extent a VLM (specifically GPT-4) is able to demonstrate a sense of ``hearing'' via being tasked with describing a perceived soundscape (including the use of onomatopoeia) from an image. We use images from 2 different sources for the following experiments. 

\subsection{Methodology}

\paragraph{Real-World}
Firstly, due to requiring high-quality publicly available images to represent the real-life condition, we utilise a subset of ImageNet.\footnote{Specifically \textit{imagenet-1k}, available at \url{https://huggingface.co/datasets/imagenet-1k}} To select our candidate images, 2 authors of this work selected a list of images that are believed to represent a wide range of soundscapes (e.g. a peaceful beach, violent waves, a desert, a car, a plane, etc.), These 50 were selected from a unique set of images that all represented different classes under the ImageNet taxonomy. Importantly, some of the chosen images were not usable with GPT-4 due to containing entities that trigger OpenAI's safeguarding restrictions (such as an image of a baby in a cot, or a couple of hunters with rifles). In such cases, we replace these images with alternative selections that the 2 authors agree are high quality. \footnote{For example, some ImageNet images are hard to decipher as they are intended to test the capabilities of computer vision models. We opt to avoid such instances and show preference towards images that present a full scene, as opposed to a single object.}

\paragraph{Generative AI}
For our GenAI-based imagery, we use scene descriptions from GPT-4 (which are generated as part of the output for this task) of the ImageNet imagery and prompt DALL-E 3 to generate images from these descriptions via the OpenAI API. In doing so, we then create a parallel dataset of 25 real-world images, 25 LLM descriptions of said images, 25 DALL-E 3 generations using the aforementioned LLM descriptions, and finally, 25 LLM descriptions of the DALL-E 3 generations.
This therefore allows us to ensure that our testing is robust to novel images, as ImageNet imagery is likely to have been a part of the training set for GPT-4's vision module. Additionally, this also facilitates an investigation as to how consistent GPT-4 is at scene description and generation (analogous to testing Neural Machine Translation via back-translation).

\paragraph{Task Setting}
We perform this task in the following way. For each condition (ImageNet/DALL-E), we present GPT-4 with the following prompt: "\textit{Imagine that the provided image is a window to another world. Describe the scene in 3 paragraphs discussing the following aspects: 
Paragraph 1: Describe what you see in the image, including the entities and the perceived environment. 
Paragraph 2: Describe what you hear in the image (i.e. the soundscape), including sounds from the identified entities, as well as the perceived environment.
Paragraph 3: In reference to the sounds mentioned in Paragraph 2, describe these sounds using onomatopoeia  (i.e. words that sound like the sounds you are trying to describe). Provide your answer to Paragraph 3 as a series of bulletpoints.}".

Following this, we ask 5 human evaluators (a subset from the main experiments), to evaluate the 3 paragraphs on a 1-5 scale, where 1 = very bad, and 5 = excellent (i.e., one rating for the visual description, one for the soundscape description, and one for the assignment of onomatopoeia to the soundscape). The instructions presented to human participants for this task are presented in Appendix \ref{apx:Materials}.\footnote{We present detailed instructions in order to moderate the understanding of what we would consider the different ratings to be indicative of in order to minimise individual perceptions of the instructions.} The order of image presentation to participants is randomised to avoid order effects.

\subsection{Results}

The results of the soundscape description task can be seen in \autoref{tab:soundscape-table} and an example generation is presented in \autoref{fig:soundscape_example}. Overall, it can be seen that human evaluators thought positively of all 3 elements asked for from GPT-4, including the visual description (which would explain performance in the following section), soundscape description and onomatopoeia, with all criteria averaging at least 4. This therefore demonstrates that GPT-4 is able to provide convincing descriptions of auditory experiences when provided with a valid image. One key thing to note is that whilst the standard deviations are consistently low, onomatopoeia demonstrates the lowest consistently. This is to be expected when evaluating a literary device, as different people may have different preferences regarding onomatopoeia they would use for certain circumstances. Additionally, there may be cases where GPT-4 has described something such as a stream and assigned the onomatopoeia ``whoosh'', which to one individual may sound too aggressive and resemble fast-moving water, when their own interpretation of a stream is more gentle (perhaps better suiting ``lap lap'').

\begin{table}
\centering \small
\begin{tabular}{lll}
\toprule
\multicolumn{3}{c}{\textbf{Soundscape Descriptions}} \\
 \midrule
 & \textbf{Mean} & \textbf{SD} \\
 \midrule
\textbf{IN Visual} & 4.19 & 0.46 \\
\textbf{IN Soundscape} & \multicolumn{1}{l}{4.42} & 0.33 \\
\textbf{IN Onomatopoeia} & 4.24 & 0.47 \\
\midrule
\textbf{D3 Visual} & 4.59 & 0.39 \\
\textbf{D3 Soundscape} & 4.55 & 0.36 \\
\textbf{D3 Onomatopoeia} & 4.24 & 0.46\\
\bottomrule
\end{tabular}
\caption{Average ratings given to the visual, soundscape, and onomatopoeia descriptions given by GPT-4 across 2 conditions. \textit{IN }refers to images from ImageNet, and \textit{D3} refers to DALL-E 3 generations.\\}
\label{tab:soundscape-table}
\end{table}

\section{GPT-4 Image ``Backtranslation''}
\label{apx:backtranslation}
We use the images and descriptions we collected to test the internal consistency of the OpenAI pipeline.\footnote{As of late November 2023.} In effect, our newly generated images demonstrate a process analogous to the back-translation used in Neural Machine Translation. To test the consistency of this process, we ask our evaluators to rate the generations on the following criteria: To what extent does the DALL-E 3 generated image present the same visual scene as the original ImageNet source image (on a scale of 1 to 5, where 1 = barely related, and 5 = all the main elements are captured).\footnote{We specify to evaluators that this task is indifferent to changes in art style, as we have not specified to DALL-E 3 that its generations should be photorealistic.}

\subsection{Results}

\autoref{tab:consistency-table} presents the results of our human evaluation. As we can see, the consistency of the pipeline is viewed favourably, with a mean rating of 4.18 across the 25 image pairings and a low standard deviation of 0.49. Importantly, no comparison was rated lower than 3 by any evaluator. This result is quite surprising given the \~100-word descriptions provided by GPT-4, indicating that GPT-4-vision is highly capable of noticing the most salient aspects of any image for recreation. The result of automatic evaluation comparing the descriptions from ImageNet and DALL-E 3 images are presented in \autoref{tab:automatic_metrics}, echoing a similar pattern to the human evaluation.

\begin{table}
\centering \small
\begin{tabular}{cc}
\toprule
\multicolumn{2}{c}{\textbf{Ratings}} \\
\midrule
\textbf{Mean} & \textbf{SD} \\
4.18 & 0.49\\
\bottomrule
\end{tabular}
\caption{Average human ratings given to the consistency of the generation pipeline when using GPT-4 descriptions of an ImageNet image to prompt DALL-E 3 to replicate. We refer to DALL-E 3 as D3 and ImageNet as IN.\\}
\label{tab:consistency-table}
\end{table}

\begin{table}
\centering \small
\begin{tabular}{ccccc}
\toprule
\multicolumn{5}{c}{\textbf{Automatic Evaluation}} \\
\midrule
\textbf{BS-P} & \textbf{BS-R} & \textbf{BS-F} & \textbf{R-L} & \textbf{BLEU} \\
\midrule
0.84 & 0.80 & 0.82 & 0.13 & 0.04\\
\bottomrule
\end{tabular}
\caption{Comparison across the visual, soundscape, and onomatopoeia descriptions from GPT-4 with the ImageNet condition as the reference and the DALL-E 3 condition as the prediction. BS-P/R/F stands for BERTScore Precision/Recall/F1, respectively. R-L is ROUGE Longest Common Subsequence. Hugging Face implementations were used for all metrics.\\}
\label{tab:automatic_metrics}
\end{table}

\subsection{Full Iconicity Prompt}
\label{apx:prompt}
\begin{figure*}
\centering
\includegraphics[width=1\linewidth]{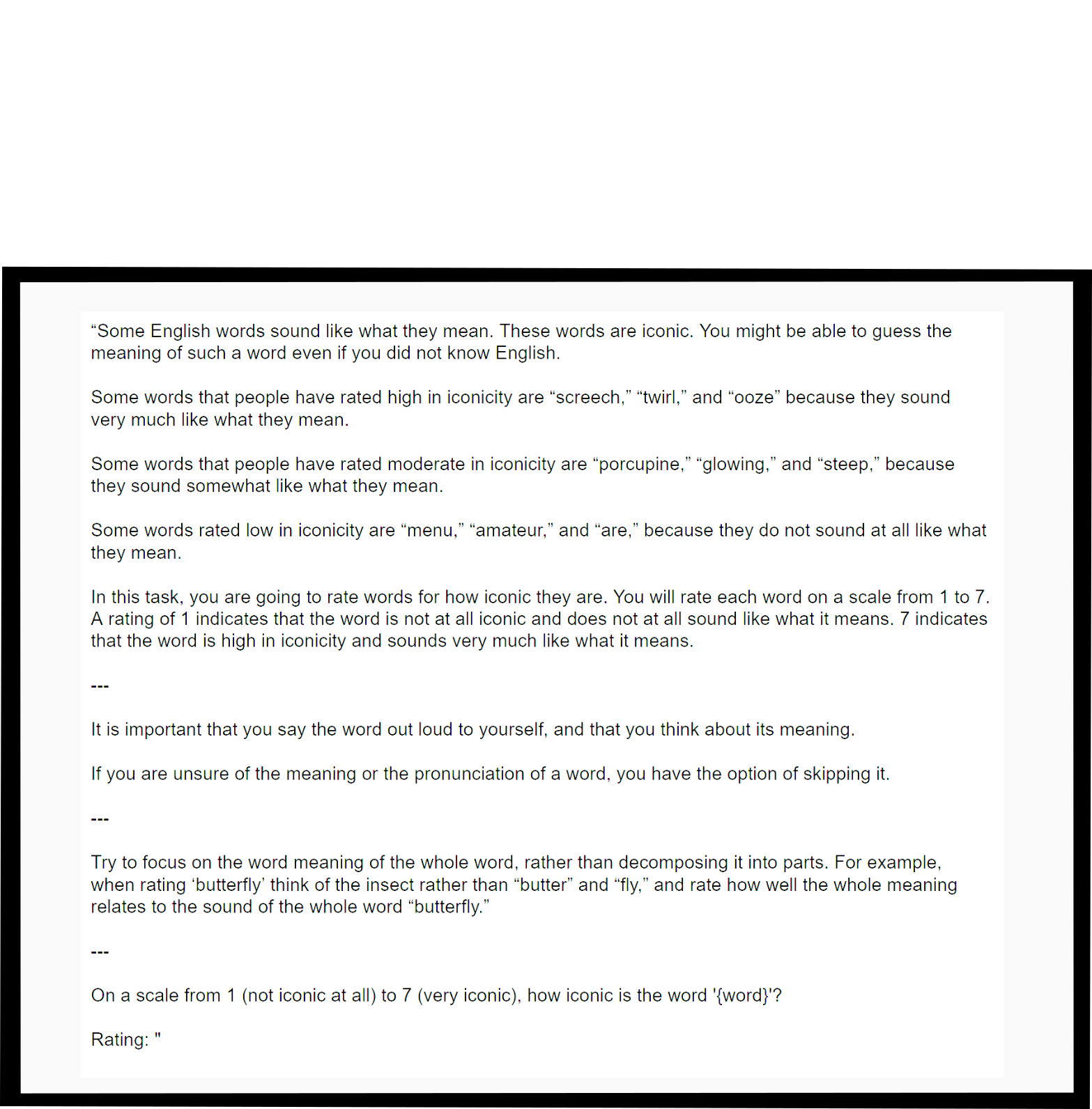}
\caption{The prompt provided to our LLMs in the iconicity judgement experiment.}
\label{fig:iconicity_prompt}
\end{figure*}

The full prompt provided to LLMs in the iconicity rating experiment is presented in \autoref{fig:iconicity_prompt}.

\subsection{Materials Provided to Participants}
\label{apx:Materials}
\begin{figure*}
\centering
\includegraphics[width=1\linewidth]{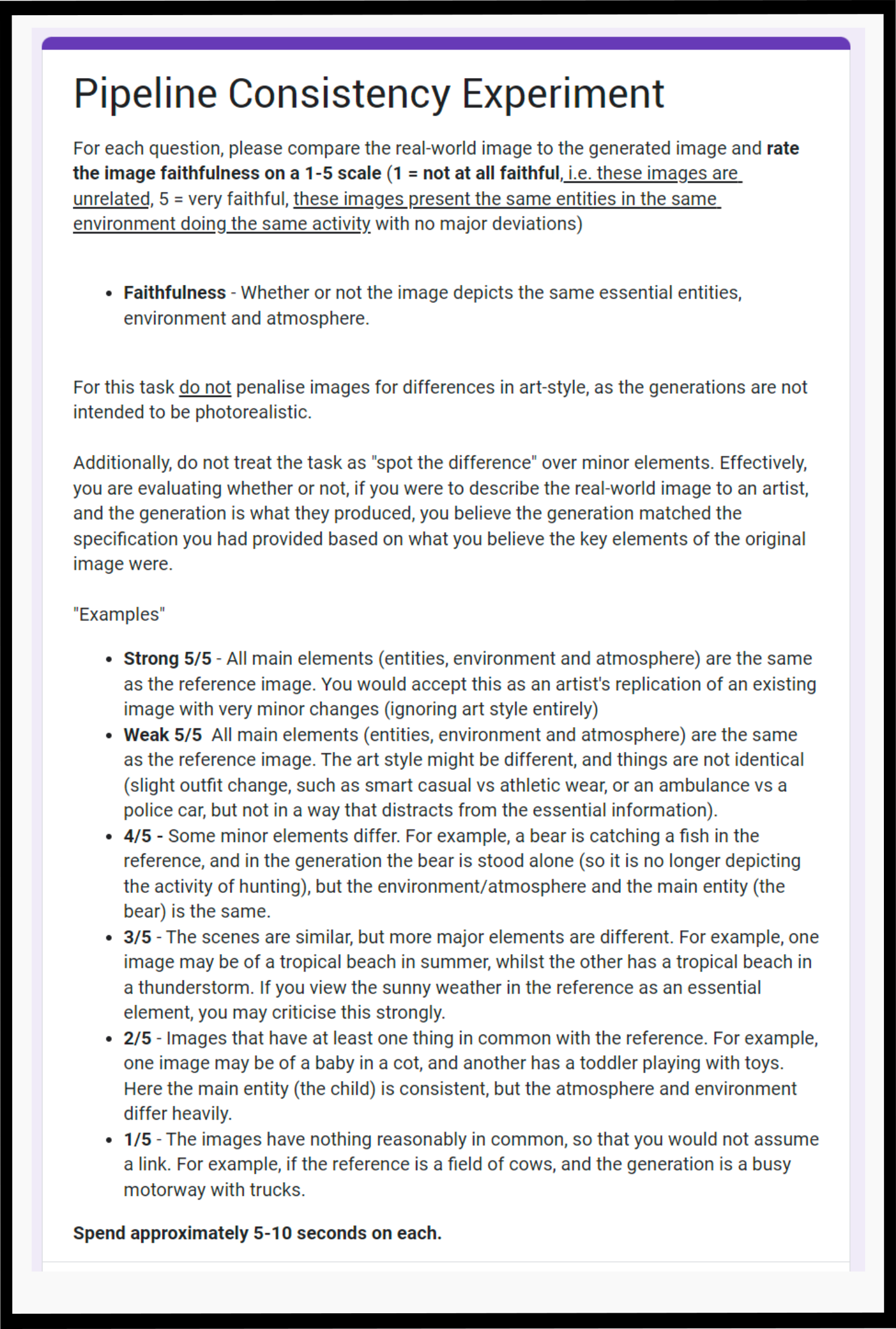}
\caption{Instructions provided to participants when asking to rate the consistency between ImageNet and DALL-E 3.}
\label{fig:comparison_instructions}
\end{figure*}

\begin{figure*}
\centering
\includegraphics[width=1\linewidth]{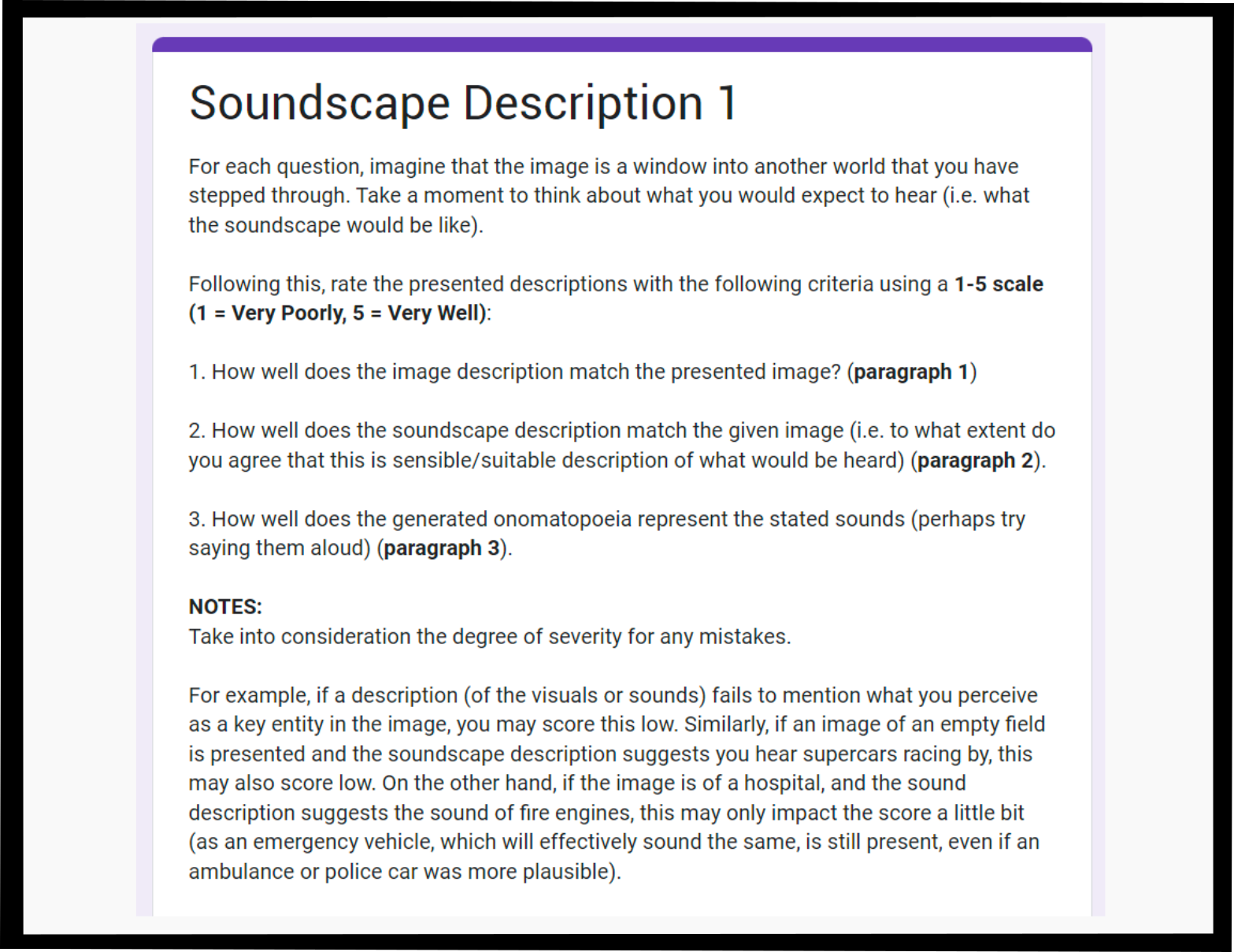}
\caption{Instructions provided to participants when asking to rate the quality of the descriptions provided by GPT-4.}
\label{fig:soundscape_instructions}
\end{figure*}

\autoref{fig:comparison_instructions} presents the instructions given to participants when rating the consistency of the OpenAI GPT-4/DALL-E 3 pipeline, whilst \autoref{fig:soundscape_instructions} presents the instructions presented to participants in the soundscape rating experiment. For Kiki-Bouba and Mil-Mal, the setup was straightforward, and participants were simply asked to select the name they believed to be the most appropriate.

\end{document}